\documentclass[runningheads]{llncs}

\usepackage{graphicx}
\usepackage{xspace}
\usepackage{amsmath}
\usepackage{mathtools}
\usepackage{amsfonts}
\usepackage{pifont}
\usepackage{dblfloatfix} 
\usepackage{floatrow}
\usepackage[dvipsnames]{xcolor}
\usepackage{subfigure}
\usepackage{booktabs}
\usepackage{pifont}
\usepackage[backref]{hyperref} 
\usepackage[figuresright]{rotating}
\usepackage{multirow}
\usepackage{tikz}
\newcommand*\circled[1]{\tikz[baseline=(char.base)]{
            \node[shape=circle,draw,inner sep=1pt] (char) {#1};}}
\newcommand{\name}[1]{\mbox{\ttfamily{#1}}\xspace}%

\newcommand{\nop}[1]{}

\usepackage{tikz}
\newcommand*{\myhash}{%
  \begin{tikzpicture}
    \pgfmathsetlengthmacro\myWidth{.8*width("=")}%
    \pgfmathsetlengthmacro\myHeight{height("H")}%
    \pgfmathsetlengthmacro\mySepY{.3333*\myWidth}%
    \pgfmathsetlengthmacro\mySideBearing{.1*\myWidth}%
    \def\myAngle{70}%
    \pgfmathsetlengthmacro\mySepX{\mySepY/sin(\myAngle)}%
    \pgfmathsetlengthmacro\mySlantX{\myHeight/tan(\myAngle)}%
    \draw[line cap=round]
      (0, {(\myHeight - \mySepY)/2}) -- ++(\myWidth, 0)
      (0, {(\myHeight + \mySepY)/2}) -- ++(\myWidth, 0)
      ({(\myWidth - \mySepX - \mySlantX)/2}, 0)
      -- ({(\myWidth - \mySepX + \mySlantX)/2}, \myHeight)
      ({(\myWidth + \mySepX - \mySlantX)/2}, 0)
      -- ({(\myWidth + \mySepX + \mySlantX)/2}, \myHeight)
    ;%
    \useasboundingbox
      (-\mySideBearing, 0)
      (\myWidth + \mySideBearing, \myHeight)
    ;%
  \end{tikzpicture}%
}

\begin{document}

\title{VisionKG: Unleashing the Power of Visual Datasets via Knowledge Graph}

\author{Jicheng Yuan\inst{1} \and
Anh Le-Tuan\inst{1} \and
Manh Nguyen-Duc\inst{1} \and
Trung-Kien Tran\inst{2} \and
Manfred Hauswirth\inst{1,3} \and
Danh Le-Phuoc\inst{1,3} 
}
\authorrunning{J. Yuan et al.}
%
\institute{Open Distributed Systems, Technical University of Berlin 
\email{\{jicheng.yuan,anh.letuan,duc.manh.nguyen, \\ manfred.hauswirth,danh.lephuoc\}@tu-berlin.de}\and
Bosch Center for Artificial Intelligence, Renningen, Germany
\email{\{TrungKien.Tran\}@de.bosch.com}\and
Fraunhofer Institute for Open Communication Systems, Berlin, Germany}
\maketitle              

\begin{abstract}
The availability of vast amounts of visual data with heterogeneous features is a key factor for developing, testing, and benchmarking of new computer vision (CV) algorithms and 
architectures. 
Most visual datasets are created and curated for specific tasks or with limited image data distribution for very specific situations, and there is no unified approach to manage and access them across diverse sources, tasks, and taxonomies. This not only creates unnecessary overheads when building robust visual recognition systems, but also introduces biases into learning systems and limits the capabilities of data-centric AI.
To address these problems, we propose the \textbf{Vision} \textbf{K}nowledge \textbf{G}raph (\textbf{VisionKG}), a novel resource that interlinks, organizes and manages visual datasets via knowledge graphs and Semantic Web technologies.  It can serve as a unified framework facilitating simple access and querying of state-of-the-art visual datasets, regardless of their heterogeneous formats and taxonomies. 
One of the key differences between our approach and existing methods is that ours is knowledge-based rather than metadata-based. It enhances the enrichment of the semantics at both image and instance levels and offers various data retrieval and exploratory services via SPARQL.
\textbf{VisionKG} currently contains  519 million RDF triples that describe approximately 40 million entities, and are accessible at {\url{https://vision.semkg.org}} and through APIs. With the integration of 30 datasets and four popular CV tasks, we demonstrate its usefulness across various scenarios when working with CV pipelines.
\end{abstract}


\section{Introduction}

Computer vision has made significant advances and visual datasets have become a crucial component in building robust visual recognition systems.
The performance of the underlying deep neural networks (DNNs) in the systems is influenced not only by advanced architectures but also significantly by the quality of training data~\cite{zhu2016we}.
There are many available visual datasets, e.g., ImageNet~\cite{deng2009imagenet}, OpenImage~\cite{kuznetsova2020open}, and MS-COCO~\cite{lin2014microsoft}, which offer a range of visual characteristics in different contexts to improve the generalization capabilities of advanced machine learning models. 
 
However, these datasets are often published in different data formats, and the quality of taxonomies and annotations varies significantly. Furthermore, labels used to define objects are available in diverse lexical definitions, such as WordNet~\cite{Miller:1995}, Freebase~\cite{Bollacker:2007}, or even just plain text. As a result, there may be inconsistencies in semantics across multiple datasets~\cite{anh:2021}. Isolated and non-unified datasets not only create unnecessary overhead when building robust visual recognition systems, but they also introduce biases into learning systems and limit the capabilities of data-centric AI~\cite{kien:2022}.

Although researchers and practitioners have made efforts to unify visual datasets~\cite{lambert2020mseg,deeplake,moore2020fiftyone}, a systematic approach to understanding the features and annotations underlying visual datasets is still lacking. 
For example, the DeepLake~\cite{deeplake} can access data from multiple data sources in a unified manner, however, it does not bridge the gap in linking and managing these datasets. Fiftyone~\cite{moore2020fiftyone} can partially capture inconsistencies in multiple datasets by visualizing data sets and analyzing data pipeline failures. Although these works improve the performance of the learned model in a data-centric manner, training DNNs with high-quality data from multiple sources in a cost-effective way remains a formidable challenge for researchers and engineers~\cite{whang2023data}.

Knowledge graph~\cite{Aidan:2021} offers a flexible and powerful way to organize and represent data that is comprehensible for both humans and machines. 
Thus, to systematically organize and manage data for computer vision, we built a knowledge graph of the visual data, named VisionKG. VisionKG is designed to provide unified and interoperable semantic representations of visual data that are used in computer vision pipelines.
This knowledge graph captures the entities, attributes, relationships, and annotations of the image data, enabling advanced mechanisms to query training data and perform further analysis. 

To address the data inconsistency problems mentioned above, VisionKG interlinks annotations across various datasets and diverse label spaces, promoting a shared semantic understanding and facilitating the retrieval of images that meet specific criteria and user requirements.
For instance, for training and testing a specific system, developers may require images with specific types and attributes tailored to their particular scenarios across a range of visual tasks or sources. For example, pedestrian and vehicle detection in adverse weather conditions~\cite{rothmeier2021performance} or occlusion-aware pose estimation~\cite{jin2022otpose} both require such tailored image sets across multiple sources for training and testing.
Our approach also enables users to better explore and understand relationships between entities using facet-based visualization and exploration powered by a graph data model. 
Graph queries powered by a graph storage can be employed to create declarative training pipelines from merged computer vision datasets, providing a convenient way to navigate and discover patterns among interlinked visual datasets such as KITTI~\cite{geiger2013vision}, MS-COCO~\cite{lin2014microsoft}, and Cityscapes~\cite{cordts2015cityscapes}. 
Additionally, VisionKG offers enhanced flexibility in terms of data representation and organization, enabling faster and easier access to the necessary information, which supports developers in building training pipelines more conveniently and efficiently.

VisionKG is built based on the Linked Data principles~\cite{bizer2011linked}, adhering to the FAIR~\cite{wilkinson2016fair} and open science guidelines~\cite{budroni2019architectures}, and  
encompasses various data sources. These sources have been defined and maintained by the research community, as they are widely used and have a significant impact on the development of computer vision algorithms and systems. Their popularity ensures that they will be regularly and frequently updated and extended. This makes VisionKG a valuable resource for researchers and developers who require access to the newest, high-quality image data. Our main contributions are summarized as follows:
\begin{itemize}
    \item We provide a unified framework for representing, querying, and analysis of visual datasets. By aligning different taxonomies, we minimize the inconsistency between different datasets.
    \item We make these datasets accessible via standardized SPARQL queries. It is available in both web user interface and via APIs.
    \item We demonstrate the advantages of VisionKG via three use cases: composing visual datasets with unified access and taxonomy through SPARQL queries, automating training and testing pipelines, and expediting the development of robust visual recognition systems.
    \item Currently, VisionKG contains 519 million RDF
triples that describe approximately 40 million entities from 30 datasets
and four popular CV tasks.
\end{itemize}

The remainder of the paper is structured as follows. In Section~\ref{Sec:VisionKG_FAIR}, we present detailed steps that follow the Linked Data publishing practice~\cite{bizer2011linked} to enforce the FAIR principles~\cite{wilkinson2016fair} in VisionKG. Section~\ref{sec:visionkg} presents the infrastructure of our VisionKG framework. In Section~\ref{Sec:VKG_MLOps}, we demonstrate the MLOps use cases with VisionKG and how it promotes this process. Sections~\ref{sec:rw}-\ref{sec:con} discuss related works and conclusions, respectively.

\section{Enforcing FAIR Principles for Visual Datasets}
\label{Sec:VisionKG_FAIR}

\subsection{Making Visual Data Assets \emph{Findable} and \emph{Accessible} } 

\begin{figure}[ht!]
    \centering
    \includegraphics[width=1\textwidth]{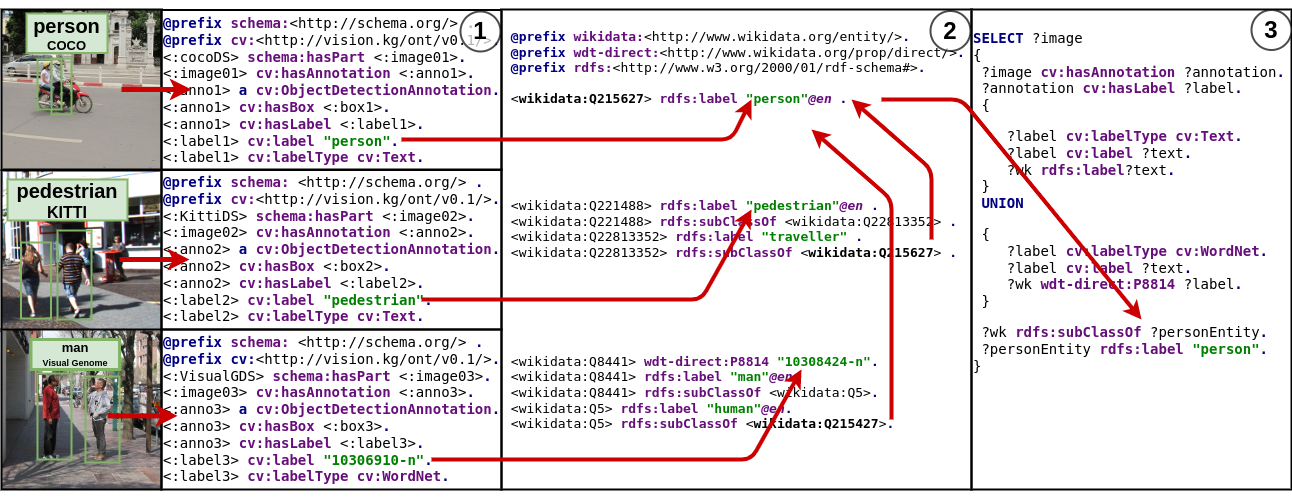}
     \vspace{-3mm}
    \caption{FAIR for Visual Data Assets}
    \label{fig01:fair}
\end{figure}

To ensure the \textbf{findability} of visual data assets, VisionKG uses Uniform Resource Identifiers (URIs) to identify resources, including images and their associated metadata. These URIs provide a unique and persistent identifier for each resource, making it easy to find and access specific images or sets of images. Figure~\ref{fig01:fair}~\circled{1} illustrates an RDF data snippet linking images and their annotations in COCO~\cite{lin2014microsoft}, KITTI~\cite{geiger2013vision} and VisualGenome~\cite{krishna2017visual}. 

This pays the way to use standardized or popular vocabularies/ontologies, such as DCAT and Schema.org to enrich metadata associated with the content and context of image data in Section~\ref{sec:inteop}. These metadata can be used to facilitate searching, filtering, and discovery of images based on specific criteria, such as object category or image resolution as demonstrated later in Section~\ref{sec:explorer}. In particular, VisionKG links each piece of metadata to a URI for the corresponding image to  ensure that metadata clearly and explicitly describe the image they refer to, e.g 'containing' bounding boxes of  'person', 'pedestrian' or a 'man' in Figure~\ref{fig01:fair}~\circled{\textbf{1}}. This  not only enables easy retrieval and exploration of images and their related ones based on their metadata but also ensures that more  metadata can be incrementally enriched by simply adding more RDF triples linked to the corresponding image. Such desired features are powered by a triple storage in terms of storing, indexing and querying (cf. Section~\ref{sec:visionkg})



In this context, VisionKG can greatly facilitate the \textbf{accessibility} of data and metadata by using standardized communication protocols and supporting the decoupling of metadata from data. Its publication practice makes it easier for targeted users to access and reuse relevant data and metadata, even when the original data are no longer available. For instance, several images of Imagenet or MSCOCO were downloaded or extracted from web sources, the metadata will provide alternative sources even the original sources are no longer accessible.

To push the \textbf{accessibility} of VisionKG's data assets even further,  users can access VisionKG through a well-documented web interface and a Python API. Both interfaces allow users to explore different aspects of VisionKG, such as the included tasks, images, and annotations with diverse semantics. Additionally, many SPARQL query examples.\footnote{\url{https://vision.semkg.org}}\textsuperscript{,}\footnote{\url{https://github.com/cqels/vision}} enable users to explore the functionalities of VisionKG in detail and describe their requirements or specific criteria using RDF statements.

\subsection{Ensure \emph{Interoperability} across Datasets and Tasks}\label{sec:inteop}

\begin{figure}[ht!]
   \centering
   \includegraphics[width=1\textwidth]{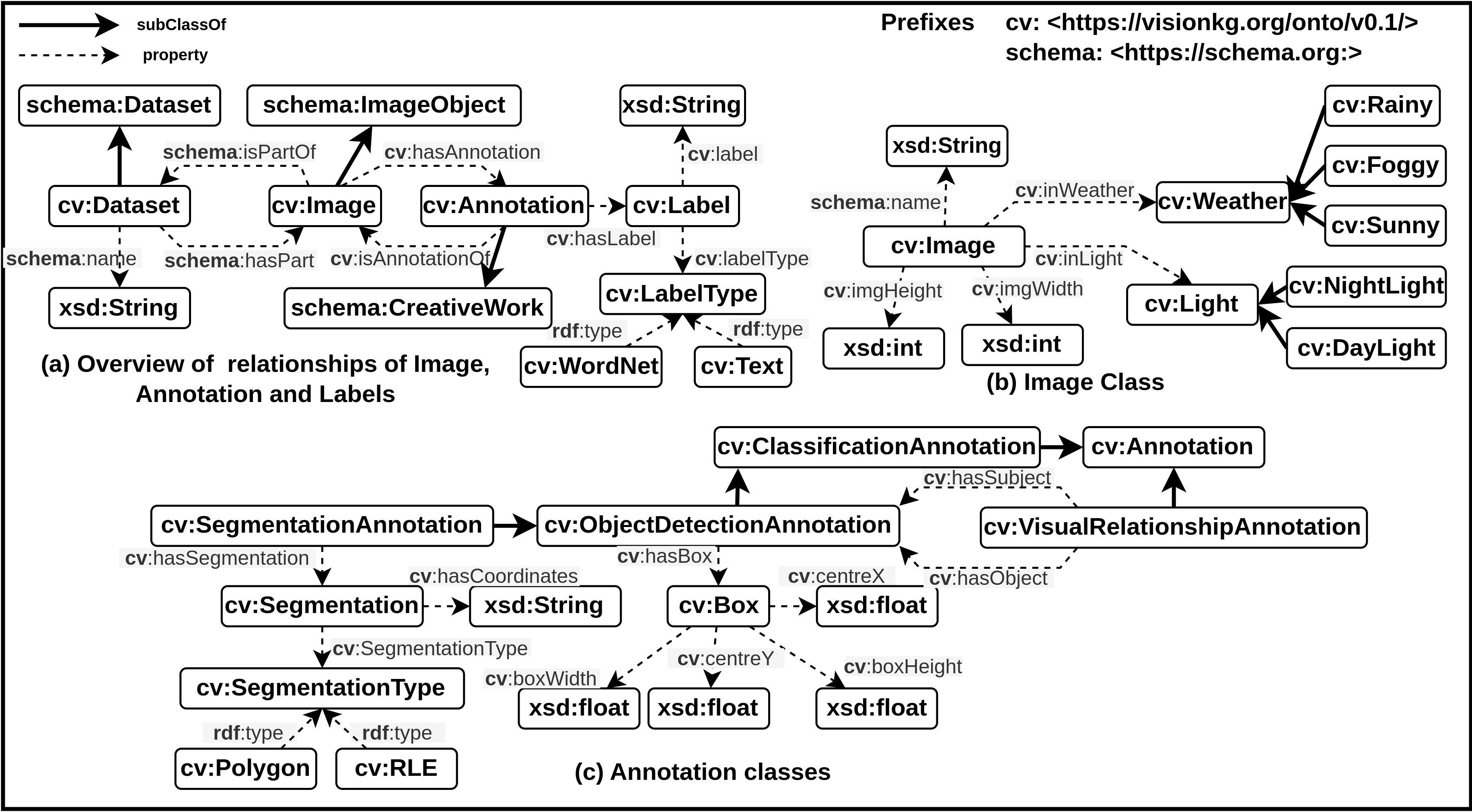}
   \caption{VisionKG Data Schema}
   \label{fig:image-overview}
\end{figure}

To make VisionKG \textbf{interoperable} across different datasets, computer vision tasks, and knowledge graph ecosystems, we designed its data schema as an RDFS ontology as shown in Figure~\ref{fig:image-overview}. This schema captures the semantics of the properties of visual data related to computer vision tasks. Our approach makes use of existing and well-developed vocabularies such as \url{schema.org} wherever possible. This ensures interoperability and backward compatibility with other systems that use these vocabularies and reduces the need for customized schema development. 

The key concepts in the CV datasets include images, annotations, and labels.
To define these concepts, we reuse the \verb|schema.org| ontology by extending its existing classes such as \verb|<schema:ImageObject>|, and \verb|<schema:CreativeWork>|.
For example, we extend \verb|<schema:ImageObject>| to create the \verb|<cv:Image>| class, \verb|<schema:Dataset>| to create the \verb|<cv:Dataset>| class. 
By doing so, we are able to inherit existing properties, such as \verb|<schema:hasPart>| or \verb|<schema:isPartOf>|, to describe the relationships between datasets and images (Figure~\ref{fig:image-overview}~\circled{a}). Our created vocabulary offers the descriptors to capture the attributes of images that are relevant for training a computer vision (CV) model (Section~\ref{sec:explorer}), such as the image dimensions, illumination conditions, or weather patterns depicted in Figure~\ref{fig:image-overview}~\circled{b}.
The concept \verb|Annotation| refers to the labeling and outlining of specific regions within an image. Each type of annotation is used for a particular computer vision task. For instance, bounding boxes are utilized to train object detection models. However, annotations are also reusable for various computer vision tasks. For example, the bounding boxes of object detection annotations can be cropped to train a classification model that doesn't require bounding boxes. In order to enable interoperability of annotations across different computer vision tasks, we developed a taxonomy for them using RDFS ontology, as illustrated in Figure~\ref{fig:image-overview}~\circled{c}. In particular, defining the object detection annotation class as a sub-class of the classification annotation enables the machine to understand that object detection annotations can be returned when users query annotations for a classification task. The cropping process can be performed during the pre-processing step of the training pipeline.

Annotations are associated with labels that define the object or relationship between two objects (visual relationship). 
However, labels are available in heterogeneous formats, and their semantics are not consistent across datasets. 
For instance, as shown in Figure~\ref{fig01:fair}~\circled{1}, the \name{pedestrian} in KITTI dataset or the \name{man} in Visual Genome dataset are annotated as \name{person} in MS-COCO dataset. Furthermore, in the Visual Genome dataset, WordNet~\cite{Miller:1995} identification is used to describe the label. Such inconsistencies make it unnecessarily challenging to combine different datasets for training or testing purposes. To tackle this issue, we assign a specific label type that indicates how to integrate with other existing knowledge graphs to facilitate the \textbf{semantic interoperability} across datasets. Figure~\ref{fig01:fair}~\circled{2} and Figure~\ref{fig01:fair}~\circled{3} exemplify how inconsistent labels from three datasets can be aligned using the RDFS taxonomies from WikiData.

\subsection{Optimize \emph{Reusability} through SPARQL Endpoint}\label{subsec:endpoint}

To optimize the reusability of visual data assets, VisionKG provides a SPARQL endpoint~\footnote{SPARQL Endpoint of VisionKG: \url{https://vision.semkg.org/sparql}} to enable users programmatically discover, combine and integrate visual data assets along with semantic-rich metadata with common vocabularies provided in Section\ref{sec:inteop}.
In particular, users can use powerful SPARQL queries to automatically retrieve desired data across datasets for various computer vision tasks. We provided exemplar queries at \url{http://vision.semkg.org/}.

Moreover, we annotated VisionKG with data usage licenses for more than ten types\footnote{List of dataset licenses in VisionKG: \url{http://vision.semkg.org/licences.html}} of licenses associated with datasets listed in Section~\ref{subsec:vkg_linkedData}. With this licensed data, users can filter datasets by their licenses to build their own custom datasets. For example, a user can pose a single SPARQL query to retrieve approximately 0.8 million training samples to train a classification model for \name{cars} with Creative Commons 4.0 license\footnote{CC BY 4.0: \url{https://creativecommons.org/licenses/by/4.0/}}.






By linking images and annotations with the original sources and related data curation processes, we captured and shared detailed provenance information for images and their annotations, thus, VisionKG enables  users to understand the history and context of data and metadata. By providing such detailed provenance information, VisionKG can enable users to better evaluate the quality and reliability of image and video data and metadata, promoting their reuse.

\section{Unified Access for Integrated Visual Datasets}
\label{sec:visionkg}

In this Section, we first provide a detailed overview of the architecture of VisionKG, and discuss how it supports access to various popular visual datasets and computer vision tasks. We then demonstrate VisionKG's capabilities in providing unified access to integrated visual datasets via SPARQL queries, ultimately promoting and accelerating data streaming in CV pipelines. 
It shows the practical usefulness of our framework for MLOps~\cite{alla2021mlops} in Section~\ref{Sec:VKG_MLOps} by exploiting knowledge graph features.

\subsection{VisionKG Architecture to Facilitate Unified Access}\label{subsec:vkg_overview}

\begin{figure}[ht!]
    \centering
    \includegraphics[width=\textwidth]{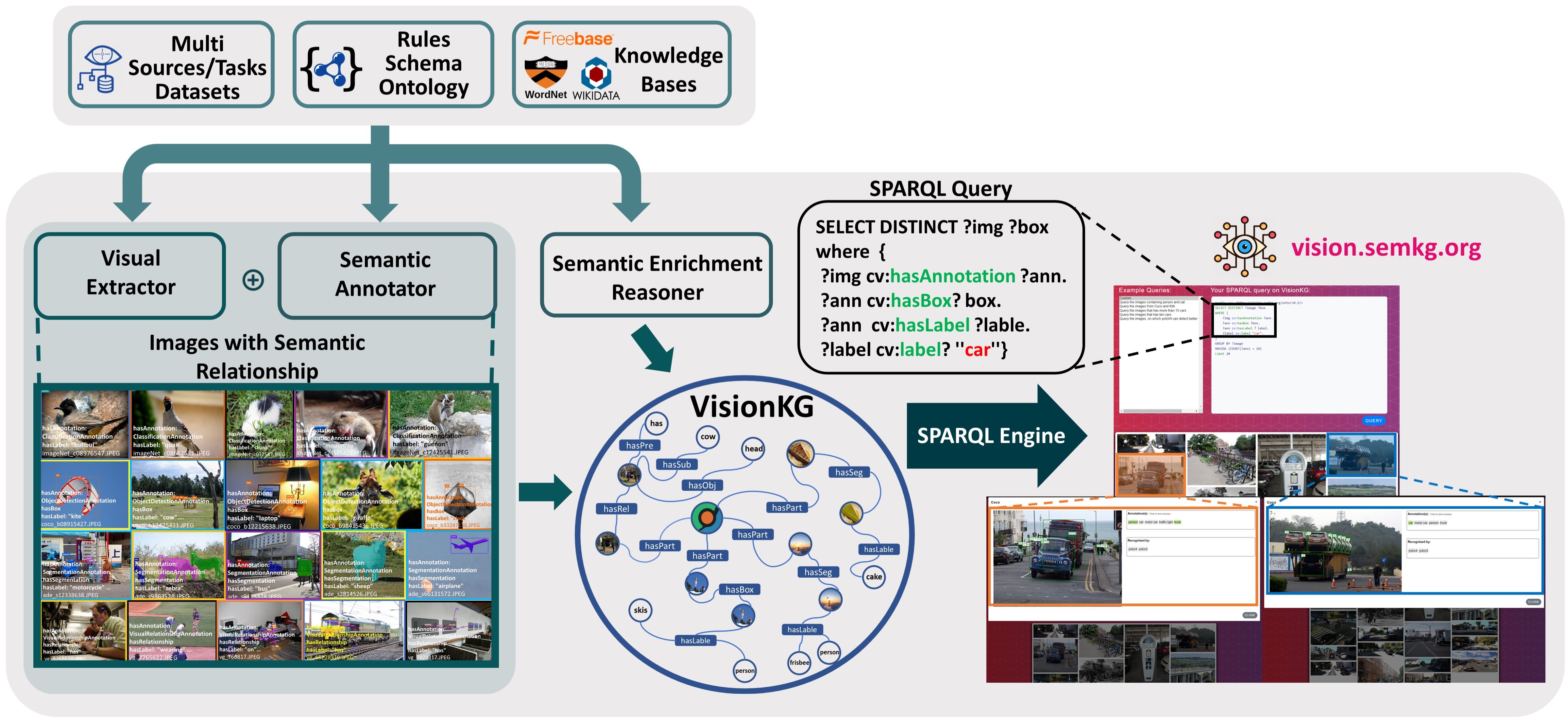}
    \caption{Overview of VisionKG Platform}
    \label{fig:visionkg-overview}
\end{figure}

Figure~\ref{fig:visionkg-overview} presents an overview of our VisionKG framework and the process of creating and enriching our unified knowledge graph for visual datasets. We start by collecting popular computer vision datasets for CV from the PaperWithCode platform~\footnote{\url{https://paperswithcode.com/datasets}}. Next, we extract their annotations and features across datasets using a \textit{Visual Extractor}. We use RDF Mapping Language (RML)\cite{Dimou:2014} to map the extracted data into RDF. RDF data is generated using a \textit{Semantic Annotator} implemented using RDFizer\cite{Iglesias:2020}. To enhance interoperability and enrich semantics in VisionKG, we link the data with multiple knowledge bases, such as WordNet~\cite{Miller:1995} and Wikidata~\cite{nielsen2018linking}. The \textit{Semantic Enrichment Reasoner} expands the taxonomy by materializing the labels in each dataset using the ontology hierarchy. For instance, categories like \name{pedestrian} or \name{man} \name{isSubClassOf} \name{person} (Figure~\ref{fig01:fair}\circled{2}). Based on the interlinked datasets and Semantic Enrichment Reasoner, users can access the data in VisionKG in a unified way (Figure\ref{fig01:fair}~\circled{3}). The SPARQL Engine maintains an endpoint for users to access VisionKG using the SPARQL query language.

Moreover, VisionKG offers a front-end web interface that allows users to explore queried datasets, such as visualizing data distribution and their corresponding annotations (\url{https://vision.semkg.org/statistics.html}).

\subsection{Linked Datasets and Tasks in VisionKG}
\label{subsec:vkg_linkedData}

The current version of our framework (by May 2023) integrates thirty most common-used and popular visual datasets, involved in the tasks for visual relationship detection, image classification, object detection, and instance segmentation. Table~\ref{tab:Statistics} gives an overview of the contained datasets, images, annotations, and triples in VisionKG. In total, it encompasses over 519 million triples distributed among these visual tasks. 

\begin{table}[ht!]
\centering
\begin{tabular}{ccccc}
\toprule[0.8pt]
\textbf{Visual Tasks}                 &\textbf{\myhash Datasets} & \textbf{\myhash Images} & \textbf{\myhash Annotations} & \textbf{\myhash Triples} \\ \hline
Visual Relationship   & 2          & 119K          &1.2M               & 2.1M              \\
Instance Segmentation & 7          & 300K          &3.9M               & 22.4M               \\
Image Classification  & 9          & 1.7M          &1.7M               & 16.6M             \\
Object Detection      & 12         & 4.3M          &50.8M               & 478.7M               \\ \hline
Total                 & 30         & 6.4M          &57.6M               & 519.8M              \\
\bottomrule[0.8pt]
\end{tabular}
\caption{Statistics across various Visual Tasks in VisionKG}
\label{tab:Statistics}
\end{table}

To enhance the effectiveness of our framework for image classification, we have integrated both large benchmark datasets, such as ImageNet~\cite{deng2009imagenet}, as well as smaller commonly used datasets, like CIFAR~\cite{krizhevsky2009learning}, the diversity of covered datasets enables users to quickly and conveniently validate model performance, thus avoid extra laborious work. Table~\ref{tab:visionkg_classification} demonstrates that ImageNet comprises 1.2 million entities, dominating the distribution of the classification task in VisionKG. Thanks to the interlinked datasets and semantic-rich relationships across visual tasks, users can query different categories and the desired number of images to tailor training pipelines for specific scenarios. 

\begin{table}[ht!]
\centering
\begin{tabular}{lccccccc}
\toprule[0.8pt]
\multicolumn{1}{c}{}           & \textbf{IMN}   & \textbf{SOP} & \textbf{CIFAR} & \textbf{MNIST}  & \textbf{CART} & \textbf{Cars196} & \textbf{CUB200} \\ \hline
\myhash  Entities & 2.7M     & 240K      &
240K & 140K    & 77K     & 32.4K     & 23.6K    \\
\myhash  Triples  & 13.3M & 1.2M   & 1.2M   & 0.7M & 0.4M  & 0.2M  & 0.1M \\ 
\bottomrule[0.8pt]
\end{tabular}
\caption{Statistics of Triples and Entities in VisionKG for Image Classification. IMN:ImageNet\cite{deng2009imagenet}, SOP: Stanford Online Products~\cite{oh2016deep}, CIFAR:CIFAR10/100~\cite{krizhevsky2009learning}, CART: Caltech-101/-256~\cite{griffin2007caltech}, CUB200: Caltech-UCSD Birds-200-2011~\cite{wah2011caltech}.}
\label{tab:visionkg_classification}
\end{table}

\begin{table}[ht!]
\centering
\begin{tabular}{lcccccccccc}
\toprule[0.8pt]
\multicolumn{1}{c}{} & \textbf{MSC}   & \textbf{UAD}  & \textbf{KIT}  & \textbf{CAR}  & \textbf{BDD}  & \textbf{OID}  & \textbf{O365} & \textbf{LVIS} & \textbf{MVD} & \textbf{VOC} \\ \hline
\myhash Entities           & 1.0M     & 678K    & 47K    & 32K    & 1.5M    & 14.3M    & 28.5M    & 1.6M    & 1.2M   & 138K   \\
\myhash Triples            & 9.7M & 6.4M & 0.4M & 0.3M & 15.1M & 135.8M & 277.4M & 15.9M    & 11.9M   & 1.0M   \\ 
\bottomrule[0.8pt]
\end{tabular}
\caption{Statistics of Triples and Entities in VisionKG for Object Detection. MSC: MS-COCO\cite{lin2014microsoft}, UAD: UA-DETRAC~\cite{wen2020ua}, KIT: KITTI~\cite{geiger2013vision}, CAR: StanfordCars196~\cite{KrauseStarkDengFei-Fei_3DRR2013}, BDD: BDD100K~\cite{yu2020bdd100k}, OID: OpenImages~\cite{kuznetsova2020open}, O365: Objects365~\cite{shao2019objects365}, LVIS~\cite{gupta2019lvis}, MVD~\cite{neuhold2017mapillary}, VOC~\cite{Everingham10}}
\label{tab:visionkg_detection}
\end{table}

For object detection, Table~\ref{tab:Statistics} and Table~\ref{tab:visionkg_detection} show that VisionKG comprises approximately 478 million triples for bounding boxes with dense annotations mainly contributed by large-scale datasets like OpenImages~\cite{kuznetsova2020open} and Objects365~\cite{shao2019objects365}. 
The variety of visual features allows users to create diverse composite datasets based on their individual requirements for the size or the density of bounding boxes, which can be helpful to reduce biases solely introduced by a single dataset captured under specific conditions and scenarios, e.g., to enhance the model performance on densely distributed small objects, which are typically challenging to localize and recognize~\cite{lin2017focal,lin2017feature}.

For visual relationship detection, which aims to recognize relationships between objects in images, we have further integrated datasets such as VisualGenome~\cite{krishna2017visual} and SpatialScene~\cite{yang2019spatialsense}, containing over 1.9 million triples for both bounding boxes and object-level relationships. Besides, VisionKG comprises 22.4 million triples for task instance segmentation, allowing users to retrieve and reuse masks of all instance-level objects for downstream scenarios, thus improving the pixel-level segmentation performance of models.

\subsection{Visual Dataset Explorer powered by SPARQL}\label{sec:explorer}

Organizing training data, which may be in heterogeneous formats and have distinct taxonomies, into one pipeline can be a time-consuming task. To reduce this effort, our framework provides a SPARQL web interface that enables users to access, explore, and efficiently combine data by leveraging the rich semantics of SPARQL. This empowers users to describe their requirements or specific criteria using graph query patterns

\begin{figure}[ht!]
   \centering
   \includegraphics[width=1\textwidth]{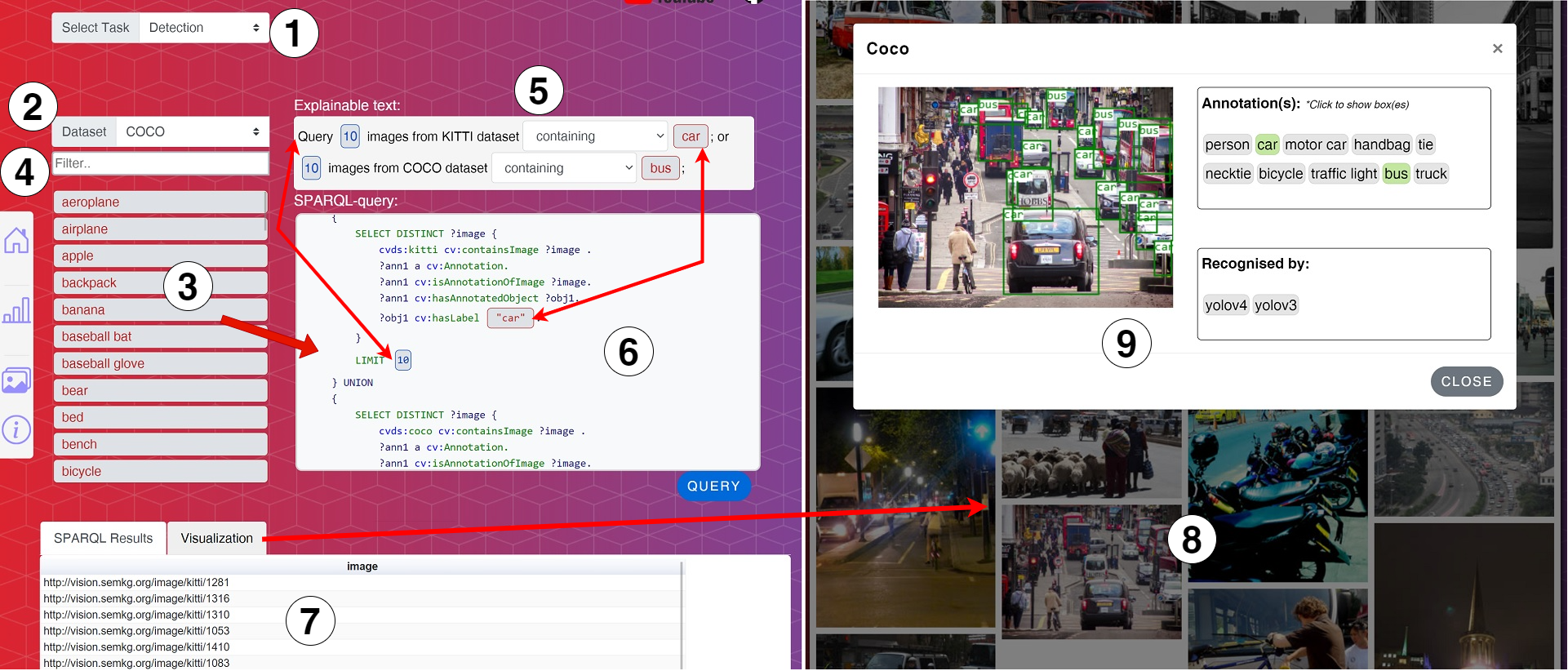}
   \caption{VisionKG Web Interface}
   \label{fig:visionkg-demo}
\end{figure}

Figure~\ref{fig:visionkg-demo} demonstrate our visual datasets explorer equipped with a live-interactive SPARQL web interface. Users can initiate their exploration by selecting a desired task, such as Detection, Classification, Segmentation, or Visual Relationship, from a drop-down menu in Figure\ref{fig:visionkg-demo}~\circled{1}. Upon task selection, the system will promptly generate a list of all compatible datasets that support the chosen task, as Figure\ref{fig:visionkg-demo}~\circled{2} illustrated.

Next, users may choose a dataset, such as COCO~\cite{lin2014microsoft} or KITTI~\cite{geiger2013vision}, from the list. This will prompt the system to display all available categories within that dataset in Figure~\ref{fig:visionkg-demo}~\circled{3}. 
To filter or select the desired categories, users can simply enter a keyword into the text box depicted in Figure~\ref{fig:visionkg-demo}~\circled{4}.
This process is further facilitated by allowing users to drag and drop a category from Figure~\ref{fig:visionkg-demo}~\circled{3} to the query box in Figure~\ref{fig:visionkg-demo}~\circled{6}.
The system will then auto-generate a SPARQL query, accompanied by an explainable text in Figure~\ref{fig:visionkg-demo}~\circled{5}, designed to select images containing the specified category. It is noteworthy that multiple categories from different datasets can be selected. Users may modify the query by removing categories or adjusting the query conditions by selecting available options from boxes in Figure~\ref{fig:visionkg-demo}~\circled{5} or Figure~\ref{fig:visionkg-demo}~\circled{6}. Additionally, users can also adjust the number of images to be retrieved.

Once the query is finalized, the user may click the "Query" button, and the results will be displayed in table format in Figure~\ref{fig:visionkg-demo}~\circled{7}. Additionally, users may select the "Visualization" tab to view the results graphically, as shown in Figure~\ref{fig:visionkg-demo}~\circled{8}. By clicking on an image, users may access additional information, such as annotations of that image and annotations generated from popular deep learning models shown in Figure~\ref{fig:visionkg-demo}~\circled{9}. Overall, the platform offers an intuitive and efficient method for dataset selection and querying for machine learning tasks.

\section{VisionKG for MLOps}
\label{Sec:VKG_MLOps}
The term \textit{MLOps} refers to the application of the DevOps workflow~\cite{ebert2016devops} specifically for machine learning (ML), where model performance is primarily influenced by the quality of the underlying data~\cite{alla2021mlops}. As demonstrated in Section~\ref{Sec:VisionKG_FAIR} and Section~\ref{sec:visionkg}, the detailed overview of our framework's architecture highlights its significant potential to boost the development of MLOps (e.g., data collection, preparation, and unified access to integrated data). In this section, we present three use cases that demonstrate how to carry out more complicated MLOps steps using our framework. These use cases demonstrate the ability to utilize VisionKG for composing visual datasets with unified access and taxonomy through SPARQL queries, automating training and testing pipelines, and expediting the development of robust visual recognition systems. VisionKG's features enable users to efficiently manage data from multiple sources, reduce overheads, and enhance the efficiency and reliability of machine learning models. More detailed features and tutorials about VisionKG can be found in our GitHub repository\footnote{\url{https://github.com/cqels/vision}}.

\subsection{Composing Visual Datasets with a Unified Taxonomy and SPARQL}
\label{Composite Visual Dataset}

Data often comes in a variety of structures and schemas, and there is a need for consolidation of this information in a unified approach. Efficient data management with expressive query functionalities plays a pivotal role in MLOps. However, data from heterogeneous sources with inconsistent formats presents numerous challenges~\cite{deeplake,renggli2021data} that must be addressed to ensure the efficiency and reliability of machine learning models under development. Additionally, the quality and consistency of data and unified access to data are paramount in developing visual recognition systems.

\begin{figure}[ht!]
    \centering
    \includegraphics[width=\textwidth]{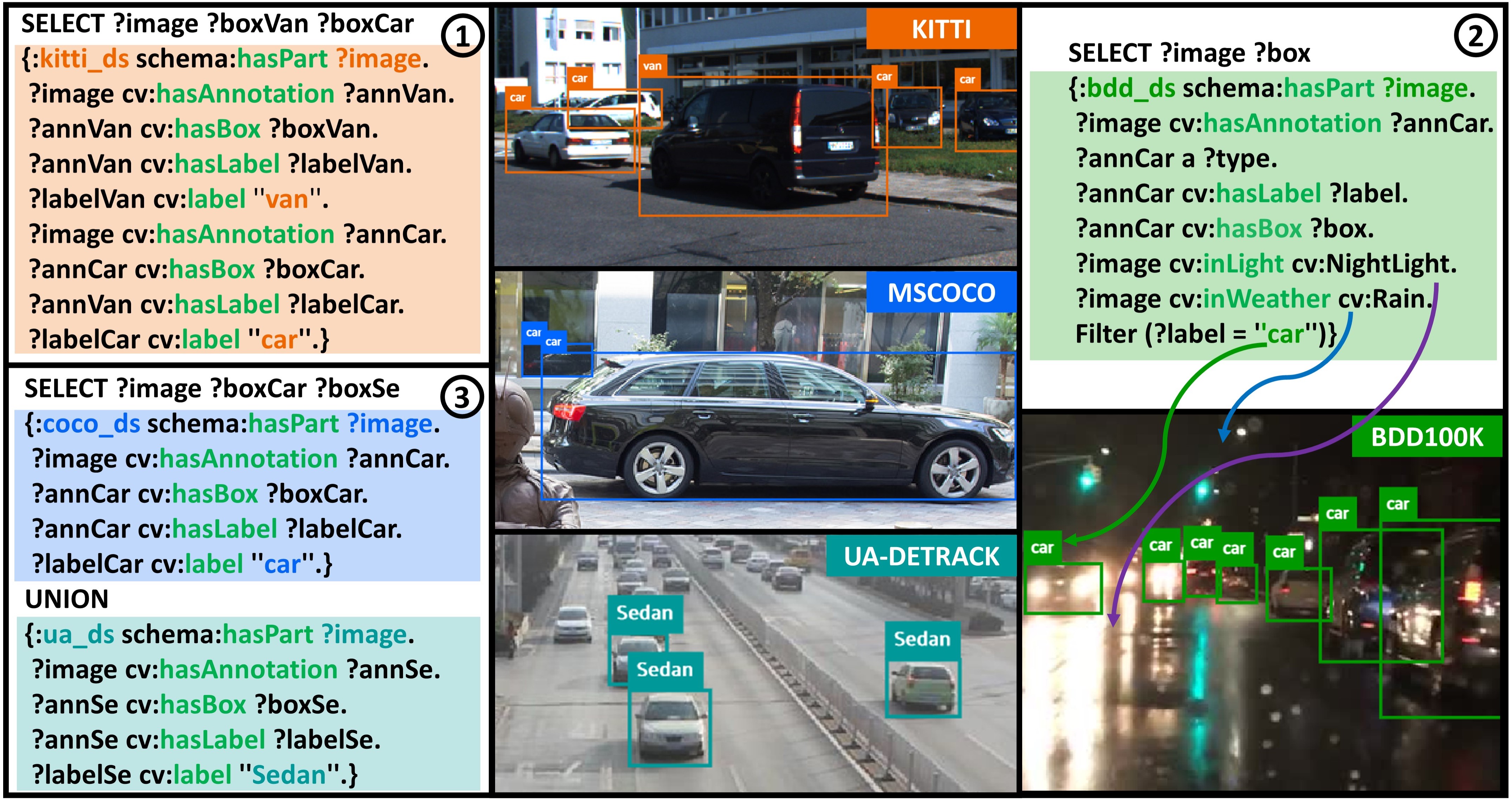}
    \caption{Dataset-Exploration with SPARQL under various Conditions in VisionKG. }
    \label{fig:UC1}
\end{figure}

As discussed in Section~\ref{Sec:VisionKG_FAIR} and ~\ref{sec:visionkg}, VisionKG is equipped with SPARQL engine allowing developers to programmatically build a composite dataset (from diverse sources with different annotated formats) to significantly reduce considerable effort in data preparation phase in MLOps. For instance, as demonstrated in Figure~\ref{fig:UC1}~\circled{1} and Figure~\ref{fig:UC1}~\circled{3}, users can query for part of images or categories from one dataset, e.g., images containing both \name{car} and \name{van} from KITTI~\cite{geiger2013vision}. Besides, as desired, they can also query for images from multiple sources with heterogeneous formats, e.g., images containing \name{car} from MS-COCO\cite{lin2014microsoft} and \name{sedan} from UA-DETRAC\cite{wen2020ua} datasets, even though they have far different annotated formats (i.e., annotations of MS-COCO and UA-DETRAC are organized in JSON and XML format, respectively). Furthermore, benefiting from the Semantic Enrichment Reasoner described in Section~\ref{subsec:vkg_overview} and integrated knowledge bases (e.g., WordNet~\cite{Miller:1995}),  users can query for images containing \name{person} from MS-COCO, KITTI, and Visual Genome~\cite{krishna2017visual} (due to distinct taxonomies, \name{person} are annotated as \name{pedestrian} in KITTI and labeled as \name{man} Visual Genome) using a simple query (Figure~\ref{fig01:fair}~\textcircled{\raisebox{-1pt}{3}}) rather than a more complex query (Figure~\ref{fig01:fair}~\textcircled{\raisebox{-1pt}{1}}) that covers all possible cases: e.g., images which containing \name{pedestrian} in KITTI and/or \name{man} in Visual Genome dataset, as users desired.

Thanks to the semantic interoperability (cf. Section~\ref{sec:inteop}) of interlinked annotations across diverse label spaces, users can create datasets from various sources with relevant definitions as desired. Along with the enrichment of semantic relationships, VisionKG provides users with composite visual datasets in a cost-efficient and data-centric manner and hence boosts the data flow in MLOps. 
Consider the Robust Vision Challenge\footnote{\url{http://www.robustvision.net/}} (RVC) in the context of object detection, where participants need to download terabyte-level datasets from the web (from different sources, taxonomies, and with inconsistent formats), and then train a unified detector to classify and localize objects across all categories in these datasets. To accomplish this, one approach is to unify the taxonomies from these datasets and mitigate the bias introduced by specific domains or similar categories (e.g., the \name{stop sign} in MS-COCO is a hyponym of \name{traffic sign}, which annotated in MVD\cite{neuhold2017mapillary}) from different taxonomies. Although the organizers provide manually aligned annotations as a good starting point, unifying labels from distinct taxonomies can still be a time-consuming process. With VisionKG, it is one step closer to achieving this. Users can carry out this process with the assistance of external knowledge bases like Wikidata~\cite{Vrandevcic:2014}, thereby leveraging external knowledge and facts. Additionally, the unified data model that leverages RDF and knowledge graphs, along with the SPARQL endpoint, allows users to conveniently query specific parts of the datasets as desired without the extra effort of parsing and processing the entire large datasets. This constitutes part of how VisionKG accelerates the MLOps workflow.

\vspace{-6mm}
\subsection{Automating Training and Testing Pipelines} 
\label{subsec:auto_training}

One of the primary goals of MLOps is to automate the training and testing pipelines to accelerate the development and deployment of ML models~\cite{alla2021mlops}. Automated workflows enable rapid iteration and experimentation, avoiding the time-consuming process during the model development for both researchers and developers. However, despite the advancements of MLOps in increased productivity and reproducibility of experiments, there are also some limitations remain in current MLOps tools (e.g., Kubeflow~\cite{bisong2019kubeflow} and MLflow~\cite{alla2021mlops} ), such as limited support for complex data types and multi-modal data (e.g., images, videos, and audios). Besides, integrating these MLOps tools with existing diverse data infrastructures can be challenging and requires significant effort.

\begin{figure}[ht!]
    \centering
    \includegraphics[width=\textwidth]{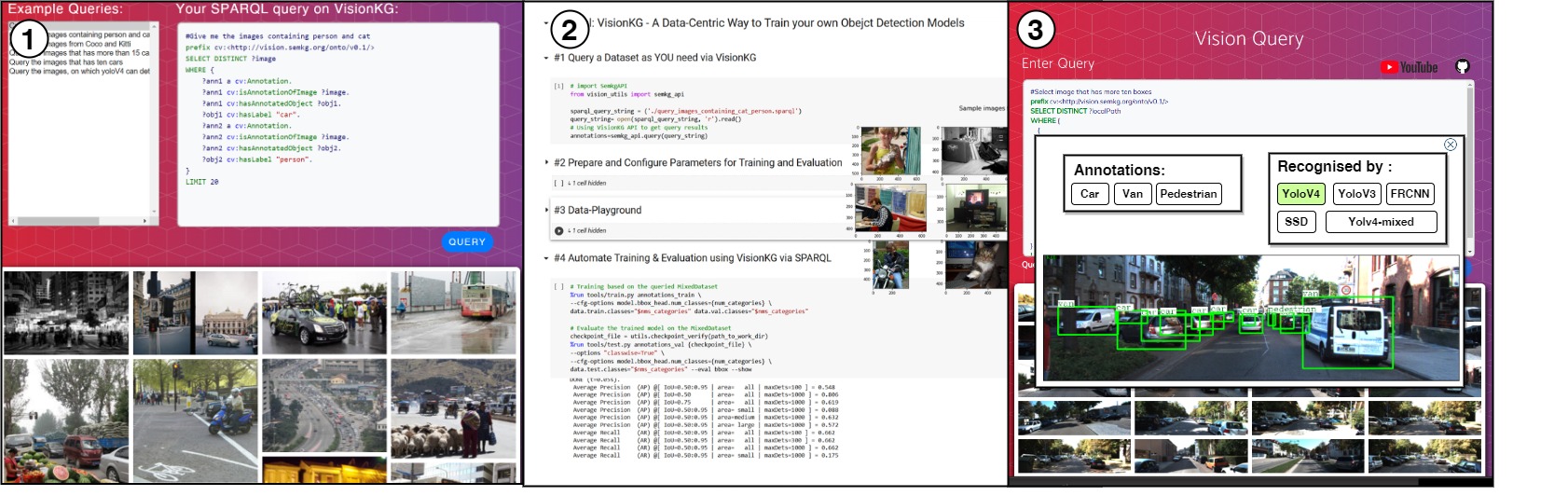}
    \vspace{-3mm}
    \caption{Construct CV Pipelines Employing VisionKG. }
    \label{fig:pipelines_from_SPARQL}
\end{figure}

As described in Section~\ref{sec:explorer}, powered by SPARQL, VisionKG supports automated end-to-end pipelines for visual tasks. Users can start a training pipeline by writing queries to construct various composite visual datasets. As demonstrated in Figure~\ref{fig:pipelines_from_SPARQL}~\circled{1}, users can query images and annotations with a few lines of SPARQL query to use RDF-based description to get desired data, such as images containing box-level annotations of \name{car} and \name{person} from interlinked datasets in VisionKG. In combination with current popular frameworks (e.g., PyTorch, TensorFlow) or toolboxes (e.g., MMDetection~\cite{chen2019mmdetection}, Detectron2~\cite{wu2019detectron2}), users can further utilize the retrieved data to construct their learning pipelines in just a few lines of Python code without extra effort, as Figure~\ref{fig:pipelines_from_SPARQL}~\circled{2} demonstrated. Users need to define solely the model they want to use and the hyperparameters they want to set.

Additionally, users can use VisionKG for their testing pipeline. The inference results can be integrated with data from VisonKG to provide quick insight about the potential model for specific scenarios. Figure~\ref{fig:pipelines_from_SPARQL}~\circled{3} demonstrates that one can gain quick overview of a trained YOLO model~\cite{jiang2022review} to detect \name{car} on images containing \name{car} in crowded traffic scenes.

\vspace{-3mm}


These described features above significantly reduce the workload during data collection, preparation, pre-processing, verification, and model selection for MLOps.
Further features of automated pipelines using VisionKG
can be found in GitHub repository\footnote{\url{https://github.com/cqels/vision}}.

\subsection{Robust Visual Learning over Diverse Data-Sources}

The increasing demand for robust visual learning systems has led to the need for efficient MLOps practices to handle large-scale heterogeneous data, maintain data quality, and ensure seamless integration between data flow and model development. Moreover, a robust learning system should perform consistently well under varying conditions, such as invariance to viewpoint and scale, stable performance under instance occlusion, and robustness to illumination changes. However, many existing visual datasets are specifically designed and curated for particular tasks, often resulting in a limited distribution of image data applicable only in narrowly defined situations~\cite{paullada2021data}. This not only imposes unnecessary burdens when developing robust visual recognition systems but also introduces biases within learning systems and constrains the robustness of visual recognition systems. 

As discussed in Section~\ref{Composite Visual Dataset} and ~\ref{subsec:auto_training}, users can use VisionKG to compose datasets across interlinked data sources and semantic-rich knowledge bases and automatically build training and testing pipelines starting from SPARQL queries. This paves the way to support the construction of robust learning systems exploiting features from VisionKG. For instance,  when users want to develop a robust object detector, besides bounding boxes and annotated categories, other environmental situations should also be considered and incorporated as prior knowledge to boost the robustness of trained detectors, such as weather and illumination conditions. Using VisionKG, as demonstrated in Figure~\ref{fig:UC1}\circled{2}, users can also employ fine-grained criteria for retrieving images with annotations, such as querying for ``images captured at \name{night} showing \name{cars} in \name{rainy} weather conditions.'' This extends VisionKG's functionalities further for exploring and constructing datasets, allowing users to explore fruitful visual features as desired and build models that cater to various scenarios more robustly, e.g., images captured during adverse weather conditions or at different times of the day. 
This potential can assist users in evaluating the capability of domain transfer of models (e.g., if a detector trained on KITTI\cite{geiger2013vision} is also robust to detect \name{cars} in \name{snowy} weather conditions) or handle rare categories and long-tail phenomenon~\cite{zhang2023deep} (e.g., query for a composite dataset containing specific categories which are rare in the source dataset to balance the data distribution).

These features reduce the bias arising from unrelated samples and also enable users to construct scenario-specific datasets covering rich semantics in a convenient fashion. In this way, it allows the users to build robust training pipelines in both data- and model-centric manners.

\vspace{-4mm}
\section{Related Work}
\label{sec:rw}

\noindent
\textbf{Limitations in Existing Computer Vision Datasets} \ \\
Modern computer vision models are data-intensive and rely heavily on available datasets to perform the learning progress and update learnable parameters. However, the majority of visual datasets are typically limited to specific domains with diverse taxonomies, and the imbalanced nature of class distribution, such as KITTI~\cite{geiger2013vision} and MS-COCO~\cite{lin2014microsoft}. Model-centric approaches, like \cite{Wang_2019_CVPR}~\cite{zhou2022simple}, have trained models to deal with those issues, they require either a domain adapter or adopt an additional model to learn the distribution of unified datasets. However, both model-centric solutions demand extra computing power. Data-centric approaches such as MSeg~\cite{lambert2020mseg} attempt to unify and interlink datasets manually which is labor-intensive. Besides, existing data toolchains or data hubs like Deep Lake~\cite{deeplake}, 
Hugging Face~\footnote{Hugging Face. \url{https://huggingface.co/docs/datasets/index}} 
and 
OpenDataLab~\footnote{Opendatalab: \url{https://github.com/opendatalab/opendatalab-python-sdk}} 
are well-established data infrastructures for organizing datasets from distinct web sources. 
However, these toolchains are based solely on meta-data and do not interlink images and annotations across datasets. In contrast, our framework employs knowledge graphs and diverse external knowledge bases to achieve this and adheres to the FAIR principles~\cite{wilkinson2016fair}, enabling VisionKG to interlink images and annotations across visual datasets and tasks with semantic-rich relationships. 
\vspace{2mm}
\\\noindent
\textbf{Knowledge Graph Technologies in Computer Vision} \ \\
Knowledge graphs can enhance the utilization of background knowledge about the real world and capture the semantic relationships in images and videos through external knowledge and facts~\cite{cui2017general,zhu2021semantic}. Approaches such as KG-CNet~\cite{fang2017object} integrate external knowledge sources like ConceptNet\cite{SpeerCH17} to capture the semantic consistency between objects in images. KG-NN~\cite{monka2021learning} facilitates the conversion of domain-agnostic knowledge, encapsulated within a knowledge graph, into a vector space representation via knowledge graph embedding algorithms. However, even these methods leverage external knowledge during learning or after the learning procedure, whereas our method utilizes not only the external knowledge bases but also interlinked datasets. In this way, the enhanced semantics can serve to render fruitful features for integrated datasets. The approach presented in \cite{filipiak2021mapping} and \cite{nielsen2018linking} use Wikidata~\cite{Vrandevcic:2014} to empower and interlink annotations for ImageNet~\cite{Krizhevsky:2012}. Thanks to the knowledge and facts from the external knowledge base, the data quality has been improved, but both are labor-intensive and  mainly target the specific dataset. Besides, the re-usability of these two approaches for other large visual datasets, such as OpenImages~\cite{kuznetsova2020open} and Objects365~\cite{shao2019objects365}, and knowledge bases, e.g., Freebase, have not been investigated. \cite{halilaj2022knowledge} employed knowledge graphs to interlink datasets. However, this approach mainly focuses on three datasets in the context of autonomous driving scenarios. In contrast, our framework, VisionKG, utilizes diverse knowledge bases such as WordNet~\cite{Miller:1995}, Wikidata~\cite{Vrandevcic:2014}, and Freebase~\cite{Bollacker:2007} to enhance the semantics in both image- and instance-level.
Additionally, KVQA~\cite{shah2019kvqa} is a knowledge-based visual dataset employing Wikidata. It is restricted mainly to \name{person} entities. Different from it, our work interlinks various visual datasets and numerous entities across diverse taxonomies and domains.

\vspace{-6mm}
\section{Conclusions and Future Works}
\label{sec:con}

We provide a novel VisionKG that serves as a unified framework for accessing and querying state-of-the-art CV datasets, regardless of heterogeneous sources and inconsistent formats.  With semantic-rich descriptions, high-quality, and consistent visual data, it not only helps to facilitate the automation of the CV pipelines but also is beneficial for building robust visual recognition systems.


As new large-scale datasets emerge, there is an increasing need to develop more efficient methods for querying and managing such a huge amount of data. As future work, we will utilize advanced indexing techniques, query optimization, and leveraging distributed computing technologies to improve scalability and integrate further datasets. 


\vspace{-6mm}
\section{Acknowledgements}
\vspace{-2mm}
This work was funded by the German Research Foundation (DFG) under the COSMO project (ref. 453130567), the German Ministry for Education and Research via The Berlin Institute for the Foundations of Learning and Data (BIFOLD, ref. 01IS18025A and ref. 01IS18037A), and the European Union's Horizon WINDERA under the grant agreement No. 101079214 (AIoTwin), and RIA research and innovation programme under the grant agreementNo. 101092908 (SmartEdge).

\newpage
\bibliographystyle{splncs04}
\bibliography{main.bib}
\end{document}